\documentclass[10pt,twocolumn,letterpaper]{article}

\usepackage{wacv24}              

\usepackage{graphicx}
\usepackage{amsmath}
\usepackage{amssymb}
\usepackage{booktabs}

\usepackage{float}
\usepackage[pagebackref,breaklinks,colorlinks]{hyperref}

\newcommand\blfootnote[1]{%
	\begingroup
	\renewcommand\thefootnote{}\footnote{#1}%
	\addtocounter{footnote}{-1}%
	\endgroup
}
\usepackage[capitalize]{cleveref}
\crefname{section}{Sec.}{Secs.}
\Crefname{section}{Section}{Sections}
\Crefname{table}{Table}{Tables}
\crefname{table}{Tab.}{Tabs.}

\usepackage{amsthm}
\usepackage{algorithm}
\usepackage{algorithmic}
\usepackage{arydshln}
\usepackage{bm}
\usepackage{color}
\usepackage{caption}
\usepackage{multirow}
\usepackage{mathtools}
\usepackage{longtable}
\usepackage{wrapfig}
\usepackage[inkscapeformat=png]{svg}
\usepackage[export]{adjustbox}

\usepackage[accsupp]{axessibility}  

\DeclareMathOperator*{\argmax}{arg\,max}
\DeclareMathOperator*{\argmin}{arg\,min}
\DeclarePairedDelimiter\ceil{\lceil}{\rceil}
\def\cT{\mathcal{T}}
\def\cY{\mathcal{Y}}
\def\cM{\mathcal{M}}

\def\btheta{\mathbf{\theta}}
\def\bx{\mathbf{x}}

\newcommand\round[1]{\left[#1\right]}
\newcommand\numberthis{\addtocounter{equation}{1}\tag{\theequation}}

\def\wacvPaperID{1868} 
\def\confName{WACV}
\def\confYear{2024}

\begin{document}

\title{Efficient Expansion and Gradient Based Task Inference for Replay Free Incremental Learning}

\author{Soumya Roy\\
Amazon\\
{\tt\small meetsoumyaroy@gmail.com}
\and
Vinay K Verma*\\
Amazon/Duke University\\
{\tt\small vinayugc@gmail.com}
\and
Deepak Gupta\\
Amazon\\
{\tt\small deepakgupta.cbs@gmail.com}
}

\maketitle

\begin{abstract}
This paper proposes a simple but highly efficient expansion-based model for continual learning. The recent feature transformation, masking and factorization-based methods are efficient, but they grow the model only over the global or shared parameter. Therefore, these approaches do not fully utilize the previously learned information because the same task-specific parameter forgets the earlier knowledge. Thus, these approaches show limited transfer learning ability. Moreover, most of these models have constant parameter growth for all tasks, irrespective of the task complexity. Our work proposes a simple filter and channel expansion-based method that grows the model over the previous task parameters and not just over the global parameter. Therefore, it fully utilizes all the previously learned information without forgetting, which results in better knowledge transfer. The growth rate in our proposed model is a function of task complexity; therefore for a simple task, the model has a smaller parameter growth while for complex tasks, the model requires more parameters to adapt to the current task. Recent expansion-based models show promising results for task incremental learning (TIL). However, for class incremental learning (CIL), prediction of task id is a crucial challenge; hence, their results degrade rapidly as the number of tasks increase. In this work, we propose a robust task prediction method that leverages entropy weighted data augmentations and the model’s gradient using pseudo labels. We evaluate our model on various datasets and architectures in the TIL, CIL and generative continual learning settings. The proposed approach shows state-of-the-art results in all these settings. Our extensive ablation studies show the efficacy of the proposed components.
\end{abstract}
\vspace{-4mm}
\section{Introduction} \blfootnote{$^\ast$The work started before joining Amazon}
Recent deep learning models outperform humans on many challenging tasks in a static environment~\cite{He2016,silver2017mastering}. However, in a continuously changing environment where novel tasks arrive sequentially, humans significantly outperform deep learning models. When presented with sequential tasks, the model suffers from catastrophic forgetting and remembers only the current task sequence. Continual learning refers to continuously learning and adapting to new environments while exploiting knowledge acquired from the past tasks without forgetting. 

The TIL and CIL are two most popular settings in continual learning. In the TIL setting, task ids are known during training and inference; however, in CIL, task ids are present only during training. The recent literature leverages three approaches to solve the continual learning problem. The replay-based methods~\cite{rebuffi2017icarl,rajasegaran2020itaml,yan2021dynamically} are the most popular approaches and show promising results, but we have to store a fraction of previous task samples for replay. Using past samples may violate privacy and increase training and storage costs. Moreover, learning is biased towards the current task since replay samples from the past tasks are limited. Regularization-based methods~\cite{Kirkpatrick2017,Aljundi2018,chaudhry2018riemannian} regularize the previous task parameters to overcome catastrophic forgetting while learning the current task. They provide sub-optimal solutions and work well only for a limited task sequence. The expansion-based methods~\cite{Rusu2016,yoon2017lifelong,zhang2019side,yoon2020scalable} are promising, as they can model many task sequences, but efficient expansion and task prediction are the critical bottlenecks. Therefore, most of the models work only for the simple TIL setting, where during inference, task ids are available. Recently~\cite{wortsman2020supermasks,singh2020calibrating,verma2021efficient} propose efficient expansion-based models, but their models do not fully utilize the previously learned parameters. They fine-tune their expansion parameter for every new task; therefore, it does not remember previous information and only the current task knowledge transfers to the next task. So, their knowledge transfer is not optimal. Moreover, most of these approaches assume that all tasks have same complexity; therefore, their expansion parameters grow equally. However, in practice, tasks are diverse and can have different complexities. Task prediction is another key challenge in expansion-based approaches; it is sensitive to the parameter and it's performance decreases rapidly as the number of tasks increase.

In this work, we propose a simple and highly efficient \emph{filter and channel} expansion-based model which is generic and can be applied to any convolutional network. The model fully utilizes the previously learned information, hence providing better knowledge transfer. Our expansion-based model learns a task-specific parameter for each novel task. However, during learning, the $i^{th}$ task model uses the global parameter and all learned task-specific parameters from $2$ to the $(i-1)^{th}$ task. Parameters up to the $(i-1)^{th}$ task preserve all the previously learned information and maximize knowledge transfer during learning of the current task. The expansion in the task-specific parameter is achieved by increasing the number of filters in each layer. We also propose a gradient-based approach to grow the task-specific parameter as a function of the task complexity; therefore for a simple task, the model has a smaller parameter growth while for complex tasks, the model requires more parameters to adapt to the current task. Our expansion-based model shows promising results in the TIL scenario. We further propose a replay-free task prediction method to solve the more challenging CIL setting. The task prediction method leverages entropy weighted augmentations and pseudo label to calculate the sample loss. Finally, we approximate the model’s gradient by the mean gradient of each layer and the norm of the task gradient is used to predict the task id.
\vspace{-0.4em}
\section{Related Work}
Recently there has been a lot of interest in the continual learning paradigm~\cite{Parisi2019}, because of its extensive applicability to various domains. We can broadly divide modern continual learning approaches into three categories - Replay, Regularization and Expansion-based models. Replay-based methods~\cite{rebuffi2017icarl,wu2018memory,castro2018end,rajasegaran2019random,Rolnick2019,rajasegaran2020itaml,van2020brain} keep a memory bank to store a fraction of samples from previous tasks. Handling sample bias~\cite{rajasegaran2020itaml} is a crucial challenge since the current task has a large number of samples, while the previous task samples are very few. Generative replay methods~\cite{Shin2017,van2018generative,liu2020generative} learn a generative model for the sample replay. 
Regularization~\cite{Kirkpatrick2017,lee2017overcoming,li2017learning,Aljundi2018} is another popular approach that regularizes the weight learned during previous tasks while training on the current task. SI~\cite{Zenke2017}, EWC~\cite{Kirkpatrick2017} and IMM~\cite{lee2017overcoming} focus on regularizing previous task weight based on their importance. These approaches cannot model a large number of tasks and model performance degrades quickly as the tasks increase. They provide sub-optimal solutions since they try to learn a joint weight that can generalize across all tasks. Recent methods like IL2A~\cite{zhu2021class}, SSRE~\cite{9878763} and FeTrIL~\cite{petit2023fetril} uses class prototypes to tackle replay-free class-incremental learning and are baselines for our method. TCL~\cite{kim2022a} provides a theoretical justification for decomposing the CIL problem into two sub-problems - TIL and task prediction. Task prediction is done using supervised contrastive learning, ensemble class prediction and output calibration by leveraging replay data.

The expansion-based models are closely related to our work, hence we mostly focus on these approaches. Recently, a wide range of methods~\cite{Rusu2016,yoon2017lifelong,xu2018reinforced,mallya2018packnet,serra2018overcoming,zhang2019side,von2019continual,gao2020efficient,yoon2020scalable,singh2020calibrating,masana2020ternary,verma2021efficient} propose expansion-based strategies to incorporate the growing task sequence. PNNs~\cite{Rusu2016} directly adds a new neural network column for each novel task. These approaches freeze the weights of the previous tasks to overcome catastrophic forgetting and lateral connections help forward knowledge transfer. Side-tuning~\cite{zhang2019side} proposes a simpler approach by training a small task-specific network and fusing the output to the base network. DEN~\cite{yoon2017lifelong} not only focuses on network expansion but also optimizes sub-problems like selective training, dynamic model expansion using loss threshold and duplication. RCL~\cite{xu2018reinforced} uses reinforcement learning to determine the growth of the architecture. Recent advancements leverage Bayesian non-parametric models where the data itself determines the expansion~\cite{Kumar2019,Lee2020,mehta2021bayesian}, but these methods work well for small datasets and few task sequences like MNIST and CIFAR-10. EFT~\cite{verma2021efficient} partitions a model into global and task-specific local parameters and leverages efficient convolution operations to construct these local transforms. The global network can be any architecture and the method outperforms most baselines on diverse task sequences in both TIL and CIL settings.

Masking~\cite{mallya2018piggyback,mallya2018packnet,serra2018overcoming,wortsman2020supermasks,masana2020ternary} is another expansion-based strategy that learns different binary/ternary masks per task. Iterative training and pruning strategies~\cite{mallya2018piggyback,hung2019compacting} have also been proposed for expansion-based continual learning. These approaches are costly to train, since pruning is an expensive step. PackNet~\cite{mallya2018packnet} requires saving masks to recover networks of previous models, which can take lots of storage space as the number of tasks grow. HAT~\cite{serra2018overcoming} proposes hard attention masks for each task. TFM~\cite{masana2020ternary} applies ternary masks to feature maps, which result in less memory per mask, as the feature maps are often smaller than the number of weights in model. TFM uses all previous tasks to learn the mask of the current task; thus, it has similar motivations as our method. APD~\cite{yoon2020scalable} decomposes the network parameters into task-shared and sparse task-specific parameters; however, the significant changes made to architecture make it harder to scale and prevent the use of pre-trained weights. SupSup~\cite{wortsman2020supermasks} finds a supermask for each task and uses gradient-based optimization to speed up inference; also, the super masks can be stored in a fixed-size Hopfield network~\cite{hopfield1982neural}. Masking approaches are promising, but they are mostly limited to the simpler TIL setting. Choosing a suitable mask requires predicting the task id, which is challenging. Our proposed approach shows promising results in both scenarios without relying on replay samples.

\vspace{-0.3em}
\section{Proposed Model}
In this section, we provide a detailed description of our proposed model. We use three key components to overcome catastrophic forgetting.
\subsection{Notations}
In incremental learning, tasks $\cT_1,\cT_2, \dots \cT_T$ arrive sequentially. Each task $\cT_i=\{(\bx_k,y_k)\}_{k=1}^{N_i}$ where $\bx_k$ has label $y_k\in\cY_i$ and $\cY_i$ is the set of all classes in task $\cT_i$ ($|\cY_i|=K_i$). During training on the $i^{th}$ task, only $\cT_i$ is available to train the model and $\cY_i\cap \cY_j=\phi$. Our proposed model can be applied to TIL as well as the CIL setting. During training and inference of the task $\cT_i$, the task id $i$ is known in the TIL setting. However, in the CIL setting, task id $i$ is unknown during inference.

\vspace{-0.1em}
\subsection{Efficient Dynamic Expansion (ablated in Sec. \ref{sec:fwdtransfer})}
    
\begin{figure}[h]
    \vspace{-3mm}
    \centering
    \includegraphics[scale=0.37]{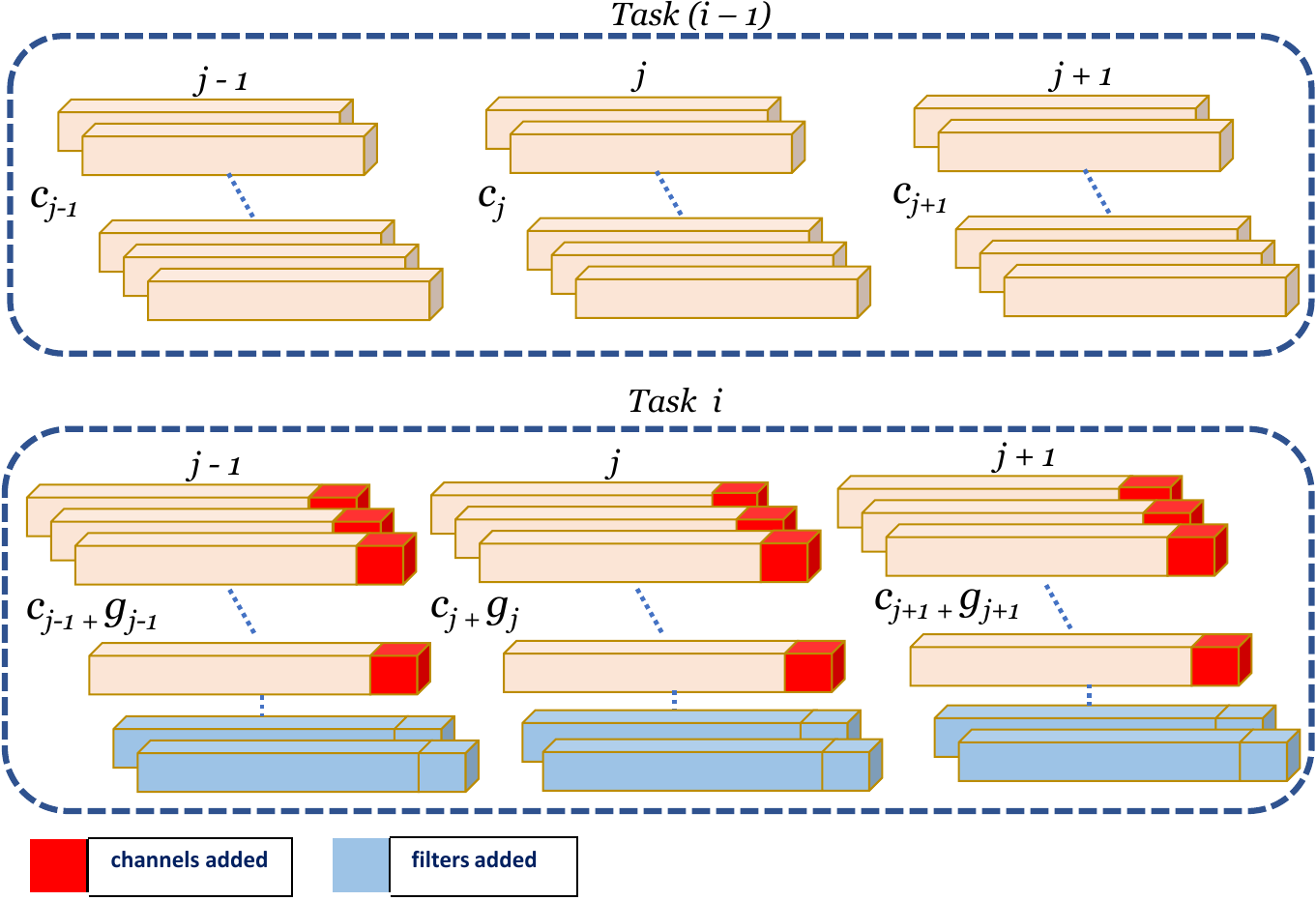}
    \caption{Expansion of the $j^{th}$ layer during task $i$ after training on the $(i-1)^{th}$ task.}
    \label{fig:expansion}
\end{figure}
In this work, we propose an efficient, dynamically expandable model which grows with the number of tasks and accumulates all the previously learned knowledge. Let $\cM_{\btheta_i}$ be the model for task $\cT_i$ where $\btheta_i$ is the model parameter. The model parameter $\btheta_i=\{\Omega,\tau_i,\tau_{2:i-1}\}$, i.e., it contains three types of parameters - global parameter ($\Omega$), task-specific parameter ($\tau_i$) and all the learned task-specific parameters before the current task ($\tau_{2:i-1}$). Therefore, the parameter $\btheta$ is growing with each novel task sequence and $|\btheta_1|<|\btheta_2|<\dots<|\btheta_T|$ where $|.|$ represents cardinality.

Now we will describe our expansion method for a particular task $\cT_i$. For task $\cT_{i-1}$, we can represent the $j^{th}$ convolutional layer for task model $\cM_{\btheta_{i-1}}$ as $l_{i-1}^{j} \in \mathbb{R}^{c_j\times k\times k\times d_j}$ where $c_j$ is the number of filters, $k$ is the kernel size and $d_j$ is the number of feature maps of layer $(j-1)$. After training on the $(i-1)^{th}$ task, we freeze it’s parameter $\btheta_{i-1}$. Let us assume that $g_{j-1}$, $g_j$ and $g_{j+1}$ are the \textit{growth rates} for the layers $(j-1)$, $j$ and $(j+1)$ respectively for the next task $\cT_i$. Before training the model for task $\cT_i$, we will grow the model by the defined growth rates and this growth is local to the $i^{th}$ task (and hence the name task-specific parameter). The $i^{th}$ task model adds $g_{j-1}$, $g_j$ and $g_{j+1}$ number of filters at the $(j-1)$, $j$ and $(j+1)$ layers respectively. Therefore, for the $i^{th}$ task at the $j^{th}$ layer, we have $c_j+g_j$ number of filters, but layer $(j-1)$ will also produce $c_{j-1}+g_{j-1}$ feature maps. To accommodate these feature maps, we have to increase the number of channels in each filter of layer $j$. So $l_{i}^{j} \in \mathbb{R}^{(c_j+g_j)\times k\times k\times d’_j}$ where $d’_j=c_{j-1}+g_{j-1}$ is the number of feature maps after adding the task-specific filters to layer $(j-1)$. Similarly, the next layer $l_{i}^{j+1} \in \mathbb{R}^{(c_{j+1}+g_{j+1})\times k\times k\times d’_{j+1}}$, where $d’_{j+1}=c_j+g_j$. Figure~\ref{fig:expansion} shows our expansion strategy at the particular layer $j$ during training of task $\cT_i$. For each task, we train batch norm and final linear layer from scratch. So the global parameter ($\Omega$) is the parameter of the first task model without batch norm and final linear layer.
\vspace{-0.2em}
\subsection{Gradient aggregation for task prediction (ablated in Sec. \ref{sec:taskpred})}
The proposed expansion-based model shows promising results in the TIL setting.
\begin{figure*}[h]
\vspace{-3mm}
    \begin{center}
    \includegraphics[scale=0.7]{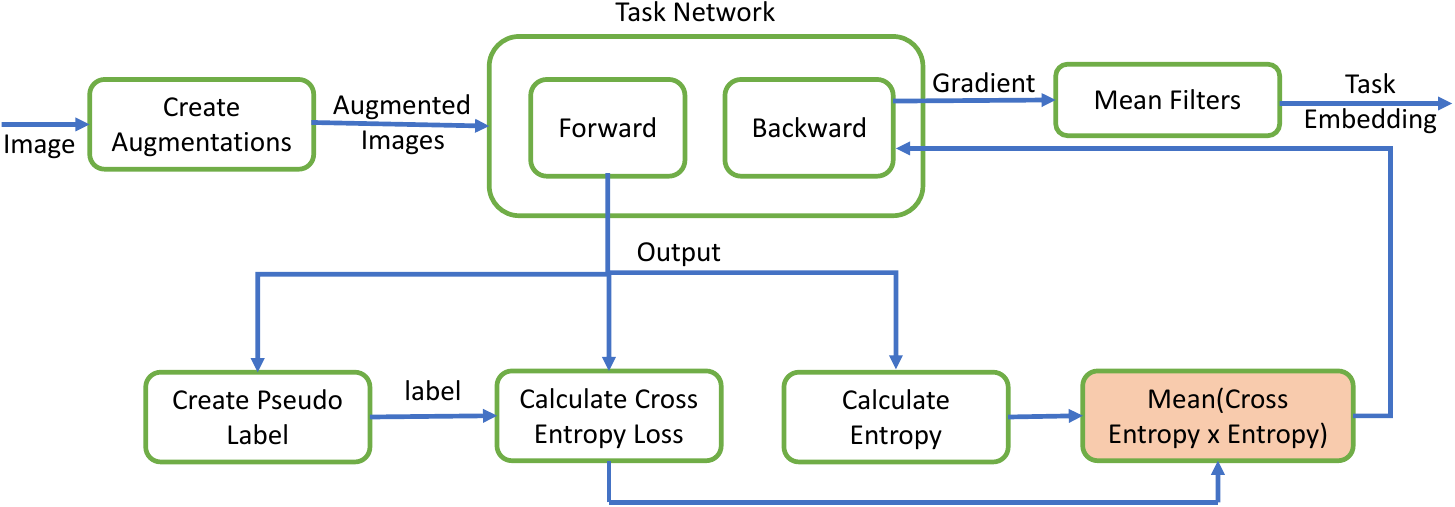}
    \end{center}
    \vspace*{-1em}
    \caption{Overview of the gradient aggregation method that leverages weighted cross-entropy using pseudo label.}
    \vspace*{-2mm}
    \label{fig:gam}
\end{figure*}
However, in the CIL setting, we don't know the task id during inference; hence we have to predict it. In this work, we propose a robust method for task prediction that can predict the task id for a large number of task sequences.

Let $\cM_{\btheta_1}, \cM_{\btheta_2}\dots \cM_{\btheta_T}$ be the learned models for the tasks $\cT_1,\cT_2, \dots \cT_T$ with parameters $\btheta_1,\btheta_2,\dots \btheta_T$. Once we have finished training for the task $\cT_i$, we can predict the sample's task id of any classes till the $i^{th}$ task. The gradient embedding of test sample $\bx_k$, for the model with parameter $\theta_i$ over the loss function $\mathcal{L}(\bx_k, \btheta_i)$, can be computed as:
\begin{equation}
    \label{eq:grad}
    \btheta'_{ik}=\nabla_{\btheta_i}\mathcal{L}(\bx_k, \btheta_i)
\end{equation}
To predict the task id of sample $\bx_k$, we measure the gradient norm using each model as follows:
\begin{equation}
    \label{eq:norm}
    \hat{i} =\argmin_{i \in T} \frac{\|\btheta_{ik}^{'}\|}{|\btheta_{ik}^{'}|}
\end{equation}
Here $|\btheta_{ik}^{’}|$ is the number of parameters in the gradient vector. We need to normalize the embedding, as each task has a different gradient length and the norm of a longer embedding can be higher. Once we have task id $\hat{i}$ for the sample $\bx_k$, we can choose the model $\cM_{\btheta_{\hat{i}}}$ to evaluate the sample. 

Minimizing the norm of loss gradient leads to flatter minima in weight space \cite{zhao2022penalizing}. A flat minima is a large connected region in weight space where the loss remains approximately constant and it is optimal minima for generalization~\cite{hochreiter1997flat,zhao2022penalizing}. In Eq.~\ref{eq:norm}, we test the shape of the loss function $\mathcal{L}$ using $\bx_k$ to estimate which model might perform better on that sample. We use $\ell_1$-norm as it gives marginally better results than $\ell_2$-norm. For a fair comparison with earlier baselines, we do not use the gradient norm loss during training time.

For a convolutional network with $L$ layers, we define the (flattened) gradient vector for layer $j$ as $\btheta_{ik}^{’j}$ ($j=1 \dots L$). An advantage of considering layer-wise gradient is that it allows us to use task and layer dependant gradient modification functions. Such functions can help us give more \textit{attention} to certain parameters in a layer. Thus we can re-write the gradient embedding for task $\cT_i$ and input $\bx_k$ as the vector:
\begin{equation}
    \label{eq:task_embed}
    \btheta_{ik}^{’}=\mathcal{CONCAT}(\mathcal{G}_{i}^{1}(\btheta_{ik}^{’1}),\mathcal{G}_{i}^{2}(\btheta_{ik}^{’2}),..,\mathcal{G}_{i}^{L}(\btheta_{ik}^{’L}))
\end{equation}
where $\mathcal{G}_{i}^{j}$ is the gradient modification function for task $i$ and layer $j$ and $\mathcal{CONCAT}$ concatenates multiple gradient vectors to create the final embedding vector. As our inference time grows linearly with the number of tasks, we can borrow the superposition idea of~\cite{wortsman2020supermasks} to get sub-linear run times. However, in this paper, we do not focus on this direction. We present an overview of our entire method in Figure~\ref{fig:gam}. Next we choose $\mathcal{L}(\bx_k, \btheta_i)$ and $\mathcal{G}_{i}^{j}$.

\vspace{-0.5em}
\subsubsection{Choosing loss function} The most common loss function for training deep networks is cross-entropy. However, task prediction using a single sample $\bx_k$ may not be robust. So our approach leverages simple test-time data augmentations to improve task prediction performance. From input $\bx_k$, we create a batch of $A$ augmented samples $\mathcal{X}_k=[\bx_{k}^1,\bx_{k}^2\dots,\bx_{k}^A]$. For each augmented sample in $\mathcal{X}_k$, we consider the class with the maximum probability as it's pseudo label and add it to the array $\hat{\mathcal{Y}_k}$. Subsequently, we take the maximally occurring pseudo label in $\hat{\mathcal{Y}_k}$ as the batch pseudo label $\hat{y}_k$. Mathematically, this can be expressed as:
\vspace{-0.5em}
\begin{equation}
    \label{eq:pseudolabel}
    \begin{aligned}
    &\hat{\mathcal{Y}_k} = [\argmax_{K_i}\btheta_i({\bx}_{k}^1),\dots,\argmax_{K_i}\btheta_i({\bx}_{k}^A)] \\
    &\hat{y}_k=\mathcal{MAXCOUNT}(\hat{\mathcal{Y}_k})
    \end{aligned}
\end{equation}
where $\mathcal{MAXCOUNT}$ takes an array as input and returns the maximally occurring element in the array. $\btheta_i({\bx}_{k}^1)$ is the output of network $\btheta_i$ for image ${\bx}_{k}^1$.

The pseudo label $\hat{y}_k$ is used to calculate the cross-entropy loss over the batch $\mathcal{X}_k$ using the model $\btheta_i$. This loss measures the robustness of the model with respect to sample perturbations. However, this augmentation strategy can create samples of different complexities, i.e., some samples can be more \textit{confusing} to the model than others. So, instead of giving equal weights to every augmented sample, we calculate the batch cross-entropy loss in a weighted manner. We leverage the entropy measure to estimate the uncertainty of each augmented sample. For augmented sample $\bx_{k}^a$ and model $\btheta_i$, we use $\mathcal{CE}(\btheta_i(\bx_{k}^a), \hat{y}_k)$ and $\mathcal{ENT}(\btheta_i(\bx_{k}^a))$ to denote the sample cross-entropy loss and sample entropy respectively. So the final loss function can be expressed as:
\vspace{-0.4em}
\begin{equation}
    \label{eq:weighted_loss}
    \mathcal{L}(\bx_k,\btheta_i)=\frac{1}{A}\sum_{a \in A} \mathcal{CE}(\btheta_i(\bx_{k}^a), \hat{y}_k) \times \mathcal{ENT}(\btheta_i(\bx_{k}^a))
\end{equation}
We use this loss to calculate the gradient in Eq.~\ref{eq:grad}. It should be noted that calculating the gradient of Eq.~\ref{eq:weighted_loss} for an augmentation $\bx_{k}^a$ is mathematically equivalent to weighing the gradients of KL-divergence and entropy terms in the corresponding cross-entropy loss $\mathcal{CE}(\btheta_i(\bx_{k}^a), \hat{y}_k)$.

\vspace{-0.8em}
\subsubsection{Choosing gradient modification function} \label{gmf}
The cardinality of the gradient in Eq.~\ref{eq:grad} is equal to the number of model parameters. Therefore, any operation on this vector is very costly. Moreover, noise in gradient directions can decrease the model's robustness. Hence, we only use the mean gradient for each convolutional filter. For fully connected layers, we use the mean gradient for each output neuron. In the standard convolutional network, the initial layers capture generic information about the input image while the later layers contain task-specific information. Hence, we use the gradients of the last two convolutional layers and the fully connected layer. We call this strategy \textit{mean filters}; it reduces the gradient cardinality by around $99.98\%$ and even helps improve task prediction.
\vspace{-0.3em}
\subsection{Static and Adaptive Growth Rate (ablated in Sec. \ref{sec:adaptgr})}
\label{grate}
Most expansion-based models~\cite{singh2020calibrating,verma2021efficient} consider the growth rate as a hyperparameter and irrespective of the task complexity, it remains constant across all tasks. Here, we propose an adaptive growth rate for the expansion parameter that depends on the task complexity. The calculation of task complexity, without availability of the previous task samples, is the key challenge. For the $i^{th}$ task, we leverage the current task samples $\cT_i$, the $(i-1)^{th}$ task model and the mean gradient on the previous task samples $\cT_{i-1}$ to construct a compatibility score. Subsequently, we use this score to decide the growth rate for the $i^{th}$ task.

Let's assume that before training for task $\cT_i$, we expand each layer $j$ of the $(i-1)^{th}$ task model by $g_j$ filters (i.e., growth rate is $g_j$). Using scaling factor $\alpha$ $\in$ [0, 1], we can write $g_j$ as:
\vspace{-0.4em}
\begin{equation}
\label{eq_ggrate}
g_j= \round{\alpha \times g_j^{min} + (1 - \alpha) \times g_j^{max}}
\end{equation}
where $g_j^{max}$ and $g_j^{min}$ are respectively the pre-defined maximum and minimum growth rates for layer $j$ and $\round{.}$ is the rounding function.

In case of static growth, the growth rate for a layer is constant for all tasks and $\alpha=0$. However, in adaptive growth, we expand our model based on the relative complexity of the novel task $\cT_i$ and use $\alpha$ to encode how \textit{similar} $\cT_i$ is with $\cT_{i-1}$. Since our models are optimized using gradient-based methods, we can measure the uncertainty of a sample using gradients~\cite{Ash2020Deep}. Given the $(i-1)^{th}$ task model, we can compare the model confidences in two samples $\bx_1$ and $\bx_2$ using $\btheta_{(i-1)1}^{'}$ and $\btheta_{(i-1)2}^{'}$ from Eq.~\ref{eq:grad}. For tasks $\cT_{i-1}$ and $\cT_i$, we can write the scaling factor $\alpha$ as:
\vspace{-0.4em}
\begin{equation}
\small
\alpha = |\mathcal{MEAN}({\{\btheta_{(i-1)k}^{'}\}}_{k \in \cT_i}) \cdot \mathcal{MEAN}({\{\btheta_{(i-1)k}^{'}\}}_{k\in \cT_{i-1}})|
\label{mean_gradient}
\end{equation}
where $\mathcal{MEAN}$ gives us the normalized mean vector from a set of input vectors. We can use this $\alpha$ to decide the growth rate for the $i^{th}$ task using Eq.~\ref{eq_ggrate}. So for each task, we only need to remember the mean gradient of the last task. We can reduce the gradient cardinality to around $0.02\%$ of it's original size using mean filters strategy.  Based on the type of growth, we get two model variants - static parameter growth (SPG) model and adaptive parameter growth (APG) model. For a fair comparison between SPG and APG models, we set $g_j^{min}$ as 1 and $g_j^{max}$ as the growth rate used in SPG. The parameter growth rate is ablated in Sec.~\ref{grate_abl}.

\vspace{-0.4em}
\section{Experiments and Results}
We conduct our experiments in TIL, CIL as well as generative continual learning settings. We also evaluate our method using different base architectures on diverse datasets. Our approach offers significant gains over the latest baselines in all the experiments.
\vspace{-0.2em}
\subsection{Datasets and Architectures}
Our experiments are performed on three datasets - CIFAR-100~\cite{Krizhevsky2009}, ImageNet-100~\cite{krizhevsky2012imagenet} and TinyImageNet~\cite{le2015tiny}. We divide the CIFAR-100 dataset into 5, 10, and 20 task sequences (let's call them CIFAR100/5, CIFAR100/10 and CIFAR100/20 respectively). CIFAR100/5 has a smaller task sequence, but each task contains 20 classes. On the other hand, CIFAR100/20 contains only 5 classes per task but has a large number of tasks. We evaluate our method on this dataset using ResNet-18~\cite{He2016} architecture for CIFAR datasets. For the ImageNet-100 dataset, we use the same class subset and class order as DER~\cite{yan2021dynamically}. Similar to CIFAR-100, we divide the ImageNet-100 dataset into 5, 10, and 20 tasks. The ImageNet-100 dataset is challenging as it contains relatively bigger images, i.e., the image size is $224\times224$. We use the standard ResNet-18 architecture for experimenting on this dataset. Tiny ImageNet is a smaller subset of the ImageNet~\cite{krizhevsky2012imagenet} dataset, contains 200 classes with a $64\times 64$ resolution. To demonstrate the efficacy of our proposed model across different architectures, we use the VGG-16~\cite{simonyan2014very} architecture with batch norm for the Tiny ImageNet dataset. We divide the dataset into 10 tasks and evaluate the model performance in the TIL setting. Furthermore, we explore our continual learning approach using the StackGAN-v2~\cite{zhang2018stackgan} architecture for three diverse tasks - cats (ImageNet~\cite{Deng2009}), birds (CUB-200~\cite{WelinderEtal2010}) and churches (LSUN~\cite{yu15lsun}).
\vspace{-0.4em}
\subsection{Baselines}
Our proposed method leverages network expansion to learn new tasks and does not use any pretrained model or sample storage for replay. Therefore, comparison with replay-based methods is not fair. So we compare our model with both regularization and expansion-based approaches. We use various regularization based methods like LwF~\cite{li2017learning}, EWC~\cite{Kirkpatrick2017}, SI~\cite{Zenke2017}, MAS~\cite{Aljundi2018}, SDC~\cite{yu2020semantic}, DMC~\cite{zhang2020class}, L2T~\cite{yoon2020scalable}, IL2A~\cite{zhu2021class}, LwF+adBiC~\cite{slim2022_transil}, SSRE~\cite{9878763} and FeTrIL~\cite{petit2023fetril} as our baselines. We also use expansion-based approaches like PNN~\cite{Rusu2016}, HAT~\cite{serra2018overcoming}, PB~\cite{mallya2018piggyback}, PackNet~\cite{mallya2018packnet}, TFM~\cite{masana2020ternary}, APD~\cite{yoon2020scalable} and EFT~\cite{verma2021efficient} for comparison. For the continual GAN experiment, we consider EWC~\cite{Kirkpatrick2017}, MeRGAN-RA~\cite{wu2018memory} and EFT~\cite{verma2021efficient} as baselines. As we share our hyperparameters with EFT~\cite{verma2021efficient}, we borrow most baseline results from that paper; the rest are obtained by either using the paper's publicly available code or by running the PyCIL~\cite{zhou2021pycil} framework for that model.
\begin{figure}[h]
\centering
    \vspace{-3mm}
    \includegraphics[scale=0.265]{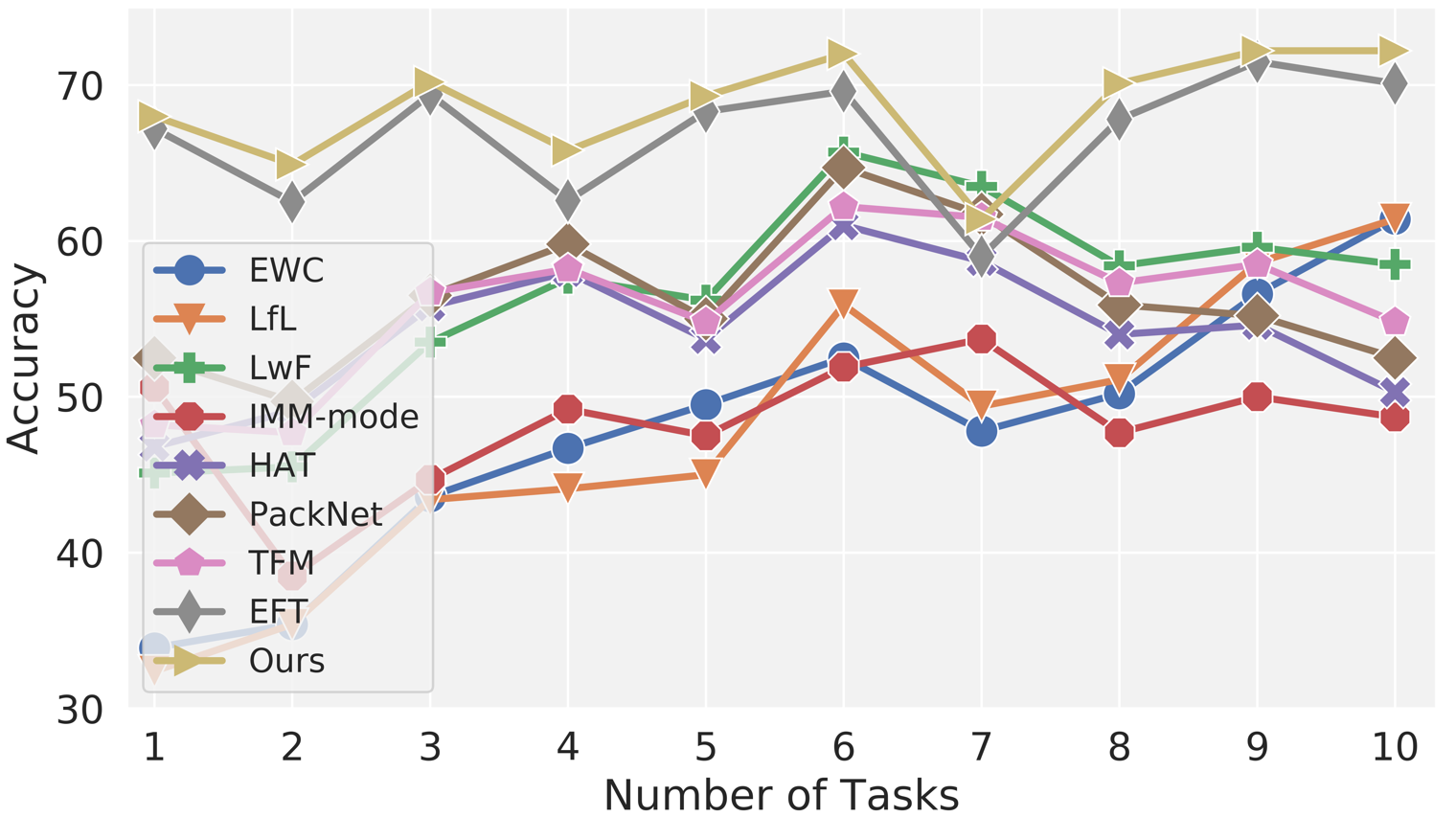}
    \vspace{-3mm}
    \caption{The comparison for the TIL scenario on TinyImageNet-200/10. The result for the $i^{th}$ task is reported after learning all the tasks.}
    \label{tab:ImageNet_tiny}
    \vspace{-6mm}
\end{figure}
\subsection{Implementation Details}
We provide the model architecture, hyperparameter settings and other experimental details for CIFAR-100, ImageNet-100 and Tiny ImageNet datasets in the supplementary material. Moreover, in supplementary, we provide results for multiple seed values along with their standard deviations.

In the following section, we discuss the results in the TIL, CIL and generative continual learning settings. We use average incremental accuracy as the evaluation metric. For TIL, we compute it as the average accuracy over all tasks, including the initial one~\cite{verma2021efficient}. In case of CIL, we report the average accuracy for all seen classes \cite{zhu2021class,verma2021efficient}.
\subsection{Results}
\subsubsection{Task Incremental Learning (TIL)}
TIL is a relatively simpler setting where the task id is known during inference. Once we know the task id, we can select the model corresponding to that id for inference. We evaluate the TIL scenario for the Tiny ImageNet~\cite{le2015tiny} dataset using the VGG-16 architecture. The Tiny ImageNet dataset contains 200 classes and we divide the dataset into 10 tasks each containing 20 classes. Our static parameter growth (SPG) model has an average parameter growth of $3.7\%$ and achieves an average incremental accuracy of $68.6\%$. In comparison, EFT~\cite{verma2021efficient} has an average incremental accuracy of $66.8\%$ with $3.6\%$ average parameter growth. On the other hand, our adaptive parameter growth (APG) model gives an average incremental accuracy of $68.5\%$ with a smaller average parameter growth of $3.4\%$. The task-wise accuracy for SPG model is shown in Fig.~\ref{tab:ImageNet_tiny}.

\subsubsection{Class Incremental Learning (CIL)}
\begin{table}[h]
\centering
    \addtolength{\tabcolsep}{-0.8mm}
	\begin{tabular}{l|l l l|l l l}
		\toprule
        & \multicolumn{3}{|c}{CIFAR-100} & \multicolumn{3}{|c}{ImageNet-100} \\
		{Methods} & {5} & {10} & {20} & {5} & {10} & {20} \\
		\midrule
		
		LwF~\cite{li2017learning} &   34.7 & 23.9 & 14.2 &  55.1 & 37.7 & 22.0\\
		EWC~\cite{Kirkpatrick2017} &  25.0 & 16.4 & 9.6 & - & - & - \\
		SI~\cite{Zenke2017}  &  29.6 & 23.3 & 13.3 & - & - & -\\
		MAS~\cite{Aljundi2018} &  24.4 & 15.4 & 10.6 & - & - & -\\
		RWalk~\cite{chaudhry2018riemannian}  &  31.6 & 17.9 & 11.0 & - & - & -\\
		EWC+SDC~\cite{yu2020semantic} & - & 19.3 & - & - & - & -\\
		DMC~\cite{zhang2020class}  & 46.6 & 36.2 & 23.9 & - & - & -\\
		IL2A~\cite{zhu2021class}&  50.2 & 29.7 & 14.7 &  40.6 & 21.7 & 9.7\\
		EFT~\cite{verma2021efficient} & 52.7 & 45.5 & 30.3 &  49.2 & 42.5 & 36.3\\
		LwF+adBiC~\cite{slim2022_transil}&  44.1 & 33.3 & 19.5 & - & - & -\\
		SSRE~\cite{9878763}&  41.9 & 30.0 & 13.7 &  37.5 & 24.0 & 16.2\\
		FeTrIL~\cite{petit2023fetril}&  45.0 & 34.3 & 20.8 &  44.0 & 31.2 & 19.8\\
		\cmidrule{1-7}
		Ours (SPG) [4.2\%] & \textbf{59.4} & \textbf{50.6} & \textbf{35.6} & \textbf{61.7} & \textbf{48.6} & \textbf{38.3}\\
		Ours (APG) [3.7\%] & \textbf{59.2} & \textbf{50.5} & \textbf{36.4} & \textbf{61.8} & \textbf{49.8} & \textbf{38.5}\\
		\bottomrule
	\end{tabular}
	\caption{Average incremental accuracy till last task for CIFAR-100 and ImageNet-100 in CIL setting. [X\%] shows the average parameter growth of the model over all task sequences and datasets.}
	\label{tab:ImageNet100c100}
\end{table}
The TIL setting requires us to know the task id during inference. This setting is not practical as, in the real word, we may not know the actual source of a datapoint. Thus, in expansion-based methods, prediction of the task id is a key challenge. To demonstrate the efficacy of our proposed gradient aggregation method for task prediction, we evaluate our model for diverse task sequences using average incremental accuracy. In all scenarios, our proposed model significantly improves over the previous state-of-the-art approaches.

In Table~\ref{tab:ImageNet100c100}, we have shown results for 5, 10 and 20 task sequences on the CIFAR-100 dataset. Using $4.3\%$ average parameter growth and static parameter growth (SPG) model variant, we achieve an absolute gain of $5.1\%$ over our best baseline EFT~\cite{verma2021efficient} in the CIFAR100/10 setting. Similarly, using $4\%$ average parameter growth, we get an absolute gain of $6.7\%$ over EFT in the CIFAR100/5 setting. The CIFAR100/20 setting is the most challenging CIL scenario as the task prediction accuracy rapidly drops with increase in task length. However, even in CIFAR100/20 setting, we achieve an absolute gain of $5.3\%$ over EFT using $4.1\%$ average parameter growth. The adaptive parameter growth (APG) model variant shows similar average incremental accuracy, but has $12.2\%$ less average parameter growth. Table~\ref{tab:ImageNet100c100} also shows results for 5, 10 and 20 task sequences on the ImageNet-100 dataset. Occasionally APG outperforms SPG as training large networks on small tasks can lead to overfitting~\cite{masana2020ternary}.

Like \cite{zhu2021class,verma2021efficient}, we report the average accuracy for all seen classes. Hence, our results for recent exemplar-free class-incremental learning (EFCIL) methods like SSRE~\cite{9878763} and FeTrIL~\cite{petit2023fetril} differ from the results reported in such papers  as they measure average accuracy for all states.
\begin{table*}[t]
 \vspace{-1.5em}
	\centering
	\addtolength{\tabcolsep}{1.8mm}
	\begin{tabular}{l | c c c  | c c c  | c c c  | l}
		\toprule
		& \multicolumn{3}{c|}{Cats} & \multicolumn{3}{c|}{Birds} & \multicolumn{3}{c|}{Churches} & Final \\
		Task $i$ & 1 & 2 & 3  & 1 & 2 & 3  & 1 & 2 & 3  & Average\\
		\midrule
		Finetune& 29.0 & 156.9 & 189.6  & - & 21.2 & 174.5 & - & - & 11.4& 125.2\\
		EWC~\cite{Kirkpatrick2017} & 29.0 & 147.3 & 190.7  & - & 65.9 & 165.4 & - & - & 38.2 & 131.4\\
		MeRGAN-RA~\cite{wu2018memory} & 29.0 & 56.4 & 58.2 & - & 50.9 & 53.7 & - & - & 23.2 &  45.1\\
		EFT~\cite{verma2021efficient} & 29.0 & 29.0 & 29.0 & - & 44.1 & 44.1 &  - & - & 32.3 & 35.1\\
		\midrule
		Ours & 29.0 & 29.0 & 29.0 & - & 40.9 & 40.9 &  - & - & 29.3 & \textbf{33.1}\\
		\bottomrule
	\end{tabular}
	\vspace{-0.5em}
	\caption{Results for GAN (StackGAN-v2) when trained in sequential manner. FID is reported after training the final task.}
	\label{tab:gan}
	\vspace{-4.9mm}
\end{table*}
\subsubsection{Generative Continual Learning}
Generative Adversarial Network (GAN)~\cite{goodfellow2014generative} also suffers from catastrophic forgetting if data arrives as a sequence of tasks. However, training a separate model for each task is costly. Hence, we apply our proposed static filter expansion approach to train the GAN architecture in a continual learning fashion. We choose the StackGAN-v2~\cite{zhang2018stackgan} architecture for the generative learning experiment the Table \ref{tab:gan} shows the results for the continual generative models.
\vspace{-0.5em}
\subsubsection{Heterogeneous Task Sequence}
\begin{table*}[!ht]
	\centering
	\addtolength{\tabcolsep}{2.2mm}
	\scalebox{0.85}{
	\begin{tabular}{c|c c|c c|c c|c c|c c|c c}
		\toprule
		{} &\multicolumn{2}{c|}{L2T} & \multicolumn{2}{c|}{PB} & \multicolumn{2}{c|}{PNN} & \multicolumn{2}{c|}{APD} &\multicolumn{2}{c|}{EFT} 
		&\multicolumn{2}{c} {Ours} \\
		\toprule
		Task order & $\downarrow$&$\uparrow$ &$\downarrow$&$\uparrow$ &$\downarrow$&$\uparrow$ &$\downarrow$&$\uparrow$&$\downarrow$&$\uparrow$&$\downarrow$&$\uparrow$\\
		\toprule
		SVHN & 10.7& 88.4& 96.8& 96.4& 96.8& 96.2& 96.8& 96.8 &96.8&95.5 & 96.6 & 95.6 \\
		CIFAR-10 &41.4& 35.8& 83.6& 90.8& 85.8& 87.7& 90.1& 91.0&89.2&90.4 & 90.5 & 91.7\\
		CIFAR-100 & 29.6& 12.2& 41.2& 67.2& 41.6& 67.2& 61.1& 67.2&64.6&71.5 & 64.5 & 71. 8\\
		\midrule
		Average &  27.2& 45.5& 73.9& 84.8& 74.7& 83.7& 83.0& 85.0 &83.5&85.8&\textbf{83.9}&\textbf{86.4}\\
		\bottomrule
	\end{tabular}
	}
	\vspace{-0.5em}
		\caption{Accuracy on the heterogeneous dataset sequence. $\downarrow$ follow the SVHN $\rightarrow$ CIFAR-10 $\rightarrow$ CIFAR-100 task order and $\uparrow$ represents CIFAR-100 $\rightarrow$ CIFAR-10 $\rightarrow$ SVHN task order.}
		\label{tab:hetero}
	\vspace{-0.8em}
\end{table*}

\begin{table}[h]
\centering
\begin{tabular}{c c c c}
		\toprule
		{Methods} & {5} & {10} & {20} \\
		\bottomrule
		EFT & 52.7 (3.9\%) & 45.5 (3.9\%) & 30.3 (3.9\%)\\
		\bottomrule
		Ours (SPG) & 59.4 (4.0\%) & 50.6 (4.3\%) & 35.6 (4.1\%) \\
		Ours (APG) & 59.2 (3.2\%) & 50.5 (3.8\%) & 36.4 (3.7\%) \\
		\bottomrule
	\end{tabular}
	\vspace{-2mm}
	\caption{Average incremental accuracy till last task with average parameter growth in brackets for CIFAR-100.}
	\label{tab:growth}
\end{table}
We evaluate our proposed static filter expansion approach on a heterogeneous task sequence. We have selected SVHN~\cite{netzer2011reading}, CIFAR-10 and CIFAR-100 datasets for constructing a task sequence and perform our experiments in the increasing (SVHN$\rightarrow$CIFAR10$\rightarrow$CIFAR100) as well as decreasing (CIFAR100$\rightarrow$CIFAR10$\rightarrow$SVHN) orders of task complexity. The task sequence is heterogeneous in two ways - firstly, SVHN and CIFAR datasets are completely different in nature and secondly, the number of classes change (increase/decrease) between tasks, i.e., go from $10\rightarrow100$ or $100\rightarrow10$. The results for the heterogeneous setting are shown in Table~\ref{tab:hetero}. We can observe that in both the scenarios (i.e., forward and reverse sequence), our approach outperforms the recent baselines using just $4.7\%$ average parameter growth.

\section{Ablations}
\subsection{Adaptive Growth Rate: Toy Experiment}
\label{sec:adaptgr}
The CIFAR-100~\cite{Krizhevsky2009} dataset contains $20$ superclasses where each superclass has $5$ similar classes. For this toy experiment, we consider $4$ superclasses - $aquatic\ mammals$, $fish$, $vehicles\ 1$ and $vehicles\ 2$. Using these superclasses, we construct two task sequences - $ordered$ and $mixed$. Each task sequence has two tasks and each task has 10 distinct classes. Task $1$ of the $ordered$ sequence contains $(\{aquatic\ mammals\}\cup\{fish\})$ classes, while task $2$ has $(\{vehicles\ 1\}\cup\{vehicles\ 2\})$ classes. On the other hand, the $mixed$ sequence picks $2-3$ classes from each superclass and constructs two tasks, each with 10 distinct classes. So the $ordered$ task sequence has two very different tasks, while the $mixed$ task sequence has two somewhat similar tasks. In this experiment, after training on the first task, we find the gradient similarity between the first and second tasks to decide the growth rate for the second task. For $ordered$ and $mixed$ task sequences, we compute the gradient similarities $\alpha_{ordered}$ and $\alpha_{mixed}$ using Eg.~\ref{mean_gradient}. We observe $\alpha_{mixed}$ $>$ $\alpha_{ordered}$ and ($\alpha_{mixed}$ - $\alpha_{ordered}$) = 0.33. This result confirms that our gradient measure can be used for finding similarities between tasks.

\begin{table}[t]
    \centering
	\addtolength{\tabcolsep}{1.1mm}
   \begin{tabular}{c c c c}
		\toprule
		{Methods} & {Growth} & {Total Params} & {Accuracy}\\
		\bottomrule
		Baseline & - & 11.2M & -\\
		\bottomrule
		EFT & 3.9\% & 15.8M & 45.5 \\
		\bottomrule
		SPG & 3.6\% & 15.8M & 51.0 \\
		SPG & 3.9\% & 16.3M & 50.3 \\
		SPG & 4.3\% & 17.0M & 50.6 \\
		\bottomrule
		APG & 3.1\% & 15.2M & 49.8 \\
		APG & 3.5\% & 15.7M & 50.1 \\
		APG & 3.8\% & 16.2M & 50.5 \\
		\bottomrule
	\end{tabular}
	\caption{Average incremental accuracy with different average parameter growths for CIFAR-100/10. Total Params is the total number of parameters that were used for training all 10 tasks.}
	\label{tab:10growth}
\end{table}

\begin{table}[t]
\vspace{-3mm}
    \centering
    \includegraphics[scale=0.055]{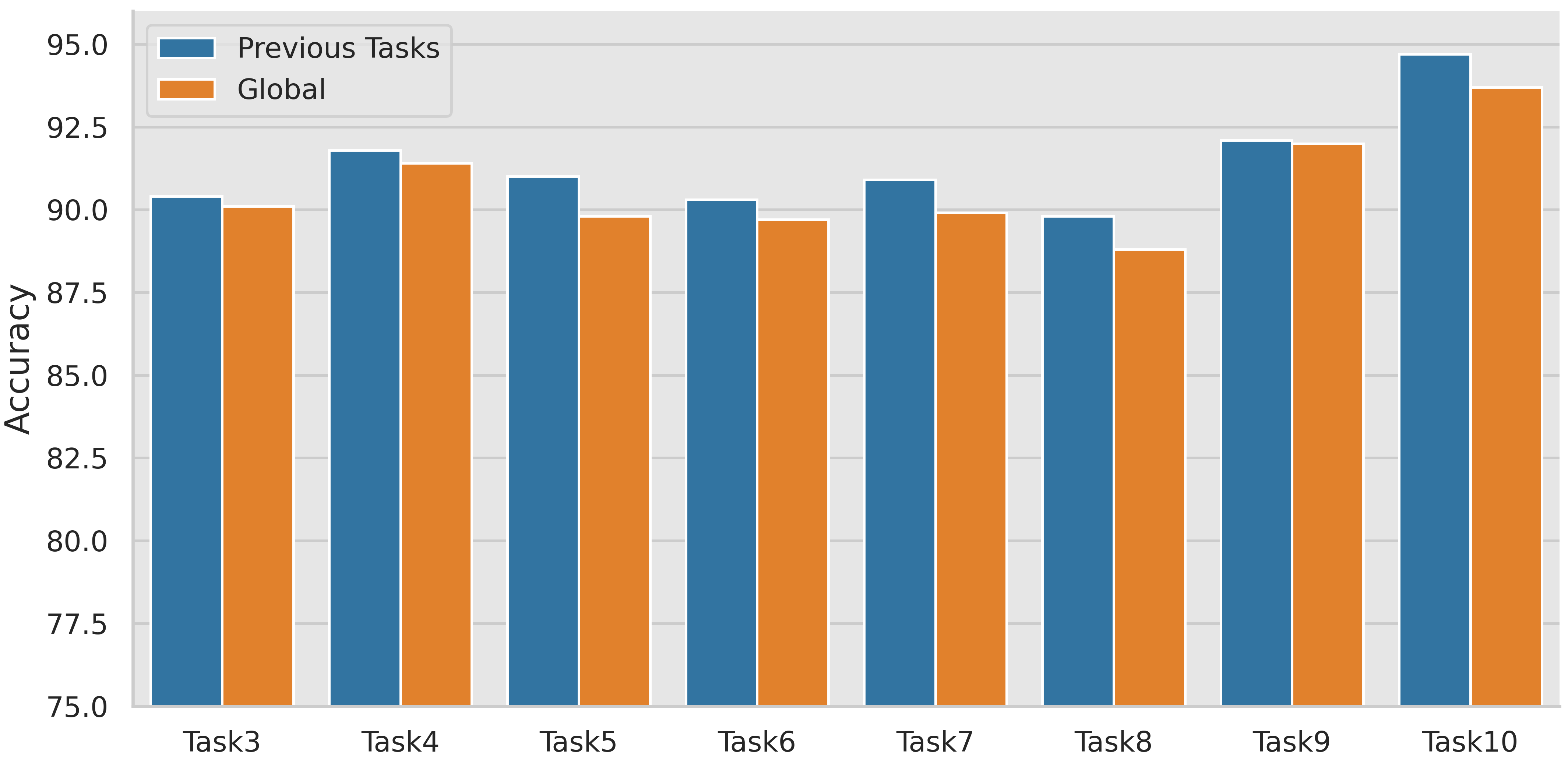}
    \vspace{-2mm}
    \captionof{figure}{The effect of forward transfer; blue is for model which leverages previous task and yellow represents model which uses global parameter.}
    \vspace{-5mm}
    \label{fig:forward_transfer}
\end{table}
\subsection{Parameter growth vs Performance}\label{grate_abl}
We define the parameter growth from task $\cT_i$ to $\cT_{i+1}$ as follows:
  $\frac{P_{i+1} - P_{i} + E_{i+1}}{P_{i}}$
where $P_{i}$ is the total number of parameters used during task $\cT_i$ and $E_{i+1}$ is the number of parameters in task $\cT_i$ which are trained from scratch during task $\cT_{i+1}$ (let's call them \textit{exclusive} parameters). For example, batch norm and the final linear layer in the ResNet-18 architecture are trained from scratch for every new task in our proposed approach and are thus counted as \textit{exclusive} parameters. We define average parameter growth for $T$ tasks as the mean of all $T$ parameter growths. Since methods like EFT~\cite{verma2021efficient} grow their networks from the first task, we define the parameter growth for the first task as the change of parameters from an equivalent standard network.

In Table~\ref{tab:growth}, we compare both our variants with our best baseline EFT~\cite{verma2021efficient} on 5, 10 and 20 splits of CIFAR-100. We show that our variants achieve significantly better results without any significant increase in average parameter growth. In Table~\ref{tab:10growth}, we show that the variants have almost no degradation in accuracy even when we significantly reduce average parameter growth. This is due to the forward transfer from previous tasks. We also show the total number of parameters that were used during training of all 10 tasks.
\vspace{-3mm}
\subsection{Forward Transfer}
\label{sec:fwdtransfer}
We have also performed experiments to demonstrate the forward transfer ability of our proposed filter expansion model. In this experiment, we have explored the scenario where our model grows over just the global or shared parameter. This type of growth is used in masking~\cite{yoon2020scalable,wortsman2020supermasks} and EFT~\cite{verma2021efficient}. We have also shown the result for our current approach where the parameter grows over the previous tasks to accumulate all the earlier knowledge. The result for the CIFAR100/10 setting is shown in Fig.~\ref{fig:forward_transfer}. Our approach shows consistently better results across all tasks.
\subsection{Task Prediction}
\label{sec:taskpred}
\begin{table}[h]
\vspace{-3mm}
	\centering
	\addtolength{\tabcolsep}{-1pt}
\begin{tabular}{p{6.5cm} |c }
		\toprule
		{Methods} & {Accuracy}  \\
		\midrule
		Ensemble Class Prediction~\cite{kim2022a} & 37.5 \\
		Entropy~\cite{verma2021efficient} & 55.3 \\
		cross-entropy & 54.9 \\
		$\nabla$(cross-entropy) + mean filters & 56.6 \\
		$\nabla$(cross-entropy) + aug + mean filters & 59.0 \\
		$\nabla$(cross-entropy) + entropy aug & 58.8 \\
		\cmidrule{1-2}
		$\nabla$(cross-entropy) + entropy aug + mean filters & \textbf{59.4} \\
		\bottomrule
	\end{tabular}
	\caption{Average incremental accuracy till last task for different task prediction methods on CIFAR100/5 split.}
	\label{tab:taskab}
	\vspace{-2mm}
\end{table}
We have experimented with different variants of our proposed task prediction method on the same task models and present the results on the CIFAR100/5 split in Table~\ref{tab:taskab}. We start off with the ensemble class prediction approach of TCL~\cite{kim2022a} and show that it gives the lowest accuracy. Subsequently, we present the result for the vanilla entropy-based task prediction approach used in multiple previous works~\cite{wortsman2020supermasks,verma2021efficient}. Lastly, we explore different components of our gradient aggregation approach and show how the accuracy improves with every step. Our proposed approach leverages the gradients of cross-entropy, entropy weighted augmentations and mean filters leading to an absolute gain of $4.1\%$ over the vanilla entropy baseline. It should be noted that entropy weighted augmentations significantly outperform vanilla augmentations for initial tasks. For example, entropy weighted augmentations lead to respective relative gains of $1.1\%$ and $1.3\%$ over vanilla augmentations for the initial $\ceil{\frac{T}{2}}$ tasks  (minus first task which has same accuracy for both baselines) on CIFAR100/5 ($T$=5) and CIFAR100/10 ($T$=10) splits.

\section{Conclusion}

In this work, we propose a highly efficient expansion-based model that accumulates all the previously learned information. The expansion rate is dynamic and it depends on the task complexity. The proposed expansion approach uses a simple and efficient in-layer \emph{filter and channel} expansion which is generic and can be applied to any convolutional network. We also propose a replay-free, flat minima based task prediction strategy which can be used to predict a large number of task sequences. The approach shows promising results for the TIL, CIL and generative continual learning settings. In expansion-based approaches, task prediction is the key limitation and it will be interesting to explore the same in future. 
\newpage

{\small
\bibliographystyle{ieee_fullname}
\bibliography{wacv}
}

\appendix
\section*{Supplementary}

In this supplementary material, we present additional ablation results and document our experimental settings.
\section{Additional Ablations}
\subsection{Influence of seeds} We use $5$ different seeds to understand how the results of our proposed method change with different seed values. The seed value impacts our method in four different ways - network initialization, sample order, random augmentation and class order. The question of seed influence is more important in adaptive parameter growth (APG) than static parameter growth (SPG) as APG uses task complexity to grow the model. As is clear from Table~\ref{tab:seeds}, our reported results are below the seed mean and reasonably stable for different seed values on all three splits of CIFAR-100. In Table~\ref{tab:seeds}, we also provide the parameter growth of a method averaged over all seeds and task sequences. It is interesting to note that the average parameter growth of the APG model is remarkably stable for different seeds. Thus, we can conclude that our methods are robust under different experimental conditions.
\begin{table*}[h]
	\centering
	\scalebox{0.967}{
		\begin{tabular}{c c c c}
			\toprule
			{Method} & {5} & {10} & {20}\\
			\midrule
			5 seeds (SPG) [4.1\%] & 59.8 $\pm$ 0.5 & 50.8 $\pm$ 0.7 & 36.9 $\pm$ 0.7 \\
			Reported (SPG) [4.1\%] & 59.4 & 50.6 & 35.6\\
			\cmidrule{1-4}
			5 seeds (APG)  [3.6\%] & 59.3 $\pm$ 0.5 & 50.8 $\pm$ 0.5 & 37.4 $\pm$ 0.7 \\
			Reported (APG) [3.6\%] & 59.2 & 50.5 & 36.4\\
			\bottomrule
		\end{tabular}
	}
	\caption{Mean and standard deviation for different splits of CIFAR-100. [X\%] shows the average parameter growth of the model over all task sequences.}
	\label{tab:seeds}
\end{table*}
\subsection{Task Prediction}
For the sake of completeness of this paper, we present the average task prediction accuracy on the CIFAR100/5 split in Table.~\ref{tab:taskab}.
\begin{table}[h]
	\centering
	\addtolength{\tabcolsep}{-1pt}
	\begin{tabular}{p{6.5cm} |c }
		\toprule
		{Methods} & {Accuracy}  \\
		\midrule
		Ensemble Class Prediction~\cite{kim2022a} & 39.8 \\
		Entropy~\cite{verma2021efficient} & 57.7 \\
		cross-entropy & 57.2 \\
		$\nabla$(cross-entropy) + mean filters & 58.8 \\
		$\nabla$(cross-entropy) + aug + mean filters & 61.3 \\
		$\nabla$(cross-entropy) + entropy aug & 61.2 \\
		\cmidrule{1-2}
		$\nabla$(cross-entropy) + entropy aug + mean filters & \textbf{61.9} \\
		\bottomrule
	\end{tabular}
	\caption{Average task prediction accuracy till last task for different task prediction methods on CIFAR100/5 split.}
	\label{tab:taskab}
	\vspace{-1em}
\end{table}
\subsection{Task-wise accuracy}
We present the task-wise accuracy of the SPG model on different splits of CIFAR-100 in Fig.~\ref{fig:c100_20}. Since this is the CIL scenario, the accuracy of a task $i$ refers to the average incremental accuracy \textit{till} task $i$. To avoid clutter, we only present results of important baselines. It should be noted that different methods have different first task accuracies as they have different optimization hyperparameters and expansion/regularization strategies. For example, EFT~\cite{verma2021efficient} expands from the first task while IL2A~\cite{zhu2021class} works better with the Adam optimizer~\cite{kingma:adam}. Similarly, the results for task-wise accuracy of the SPG model on different splits of ImageNet-100 is shown in Fig.~\ref{fig:im100_20}.
\begin{figure*}
	\setlength{\lineskip}{0pt}
	\centering
	\begin{subfigure}[H]{1\textwidth}
		\centering
		\includegraphics[width=0.8\textwidth]{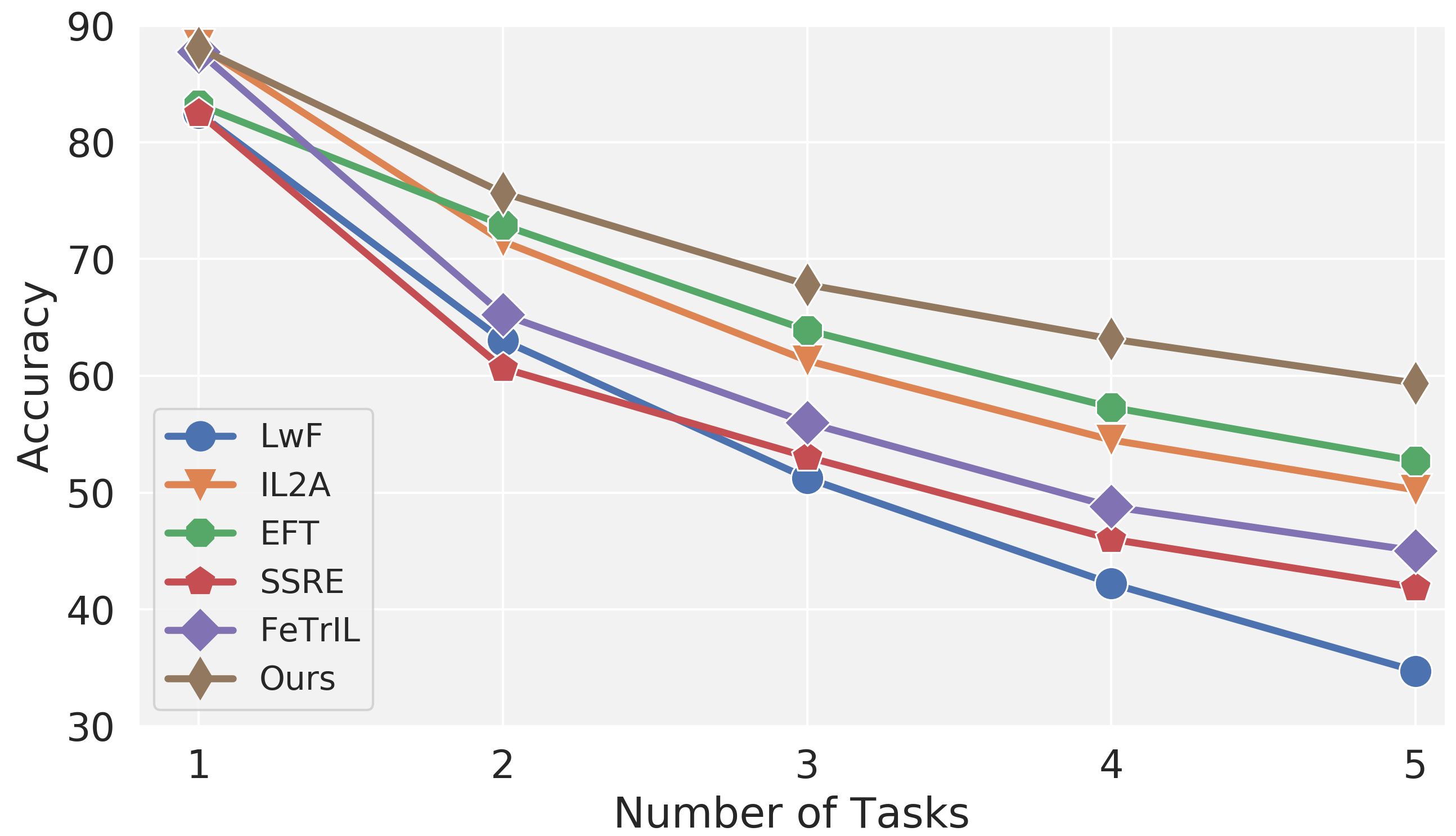}
	\end{subfigure} \hfill
	\begin{subfigure}[H]{1\textwidth}
		\centering
		\includegraphics[width=0.8\textwidth]{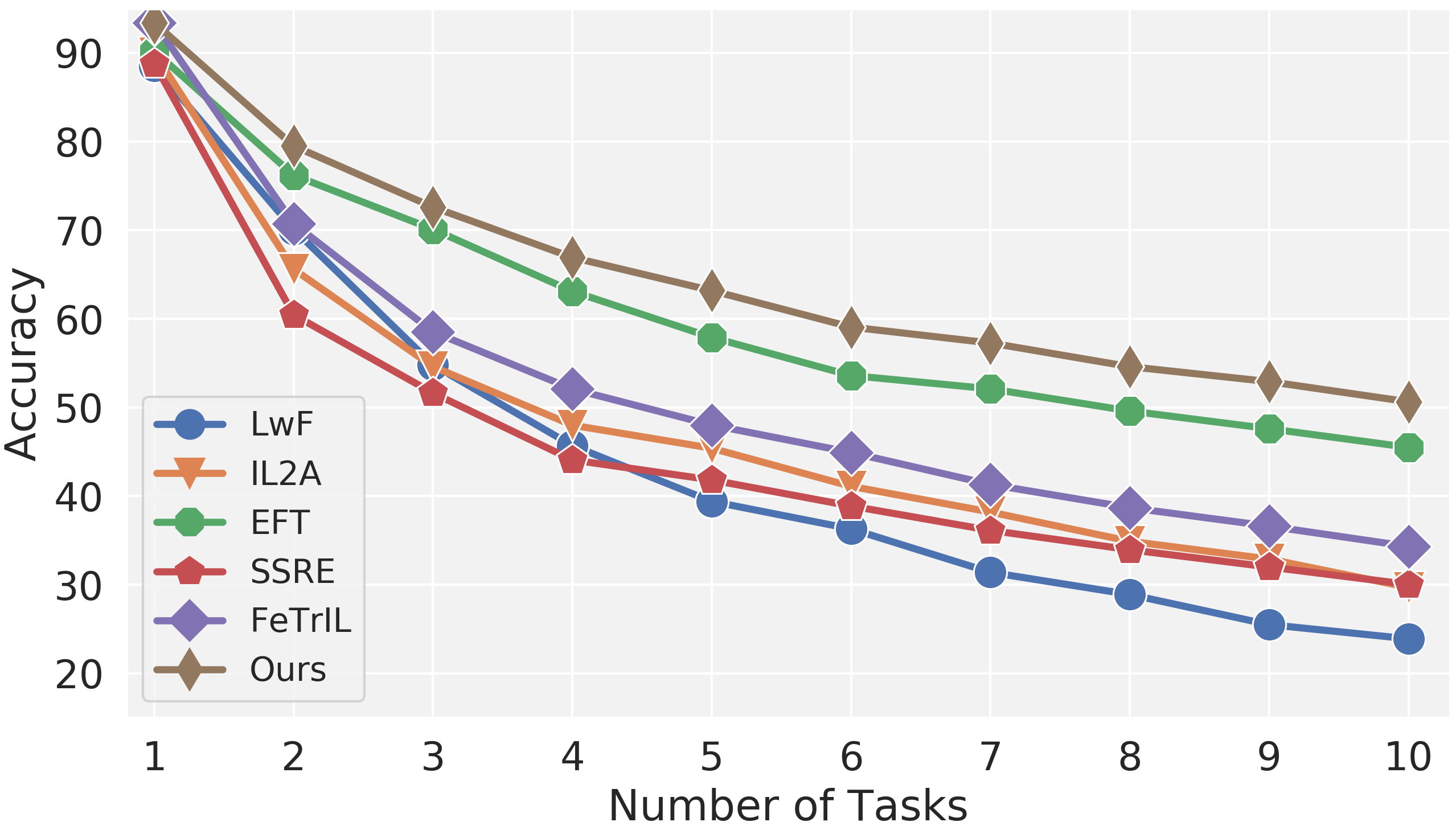}
	\end{subfigure} \hfill
	\begin{subfigure}[H]{1\textwidth}
		\centering
		\includegraphics[width=0.8\textwidth]{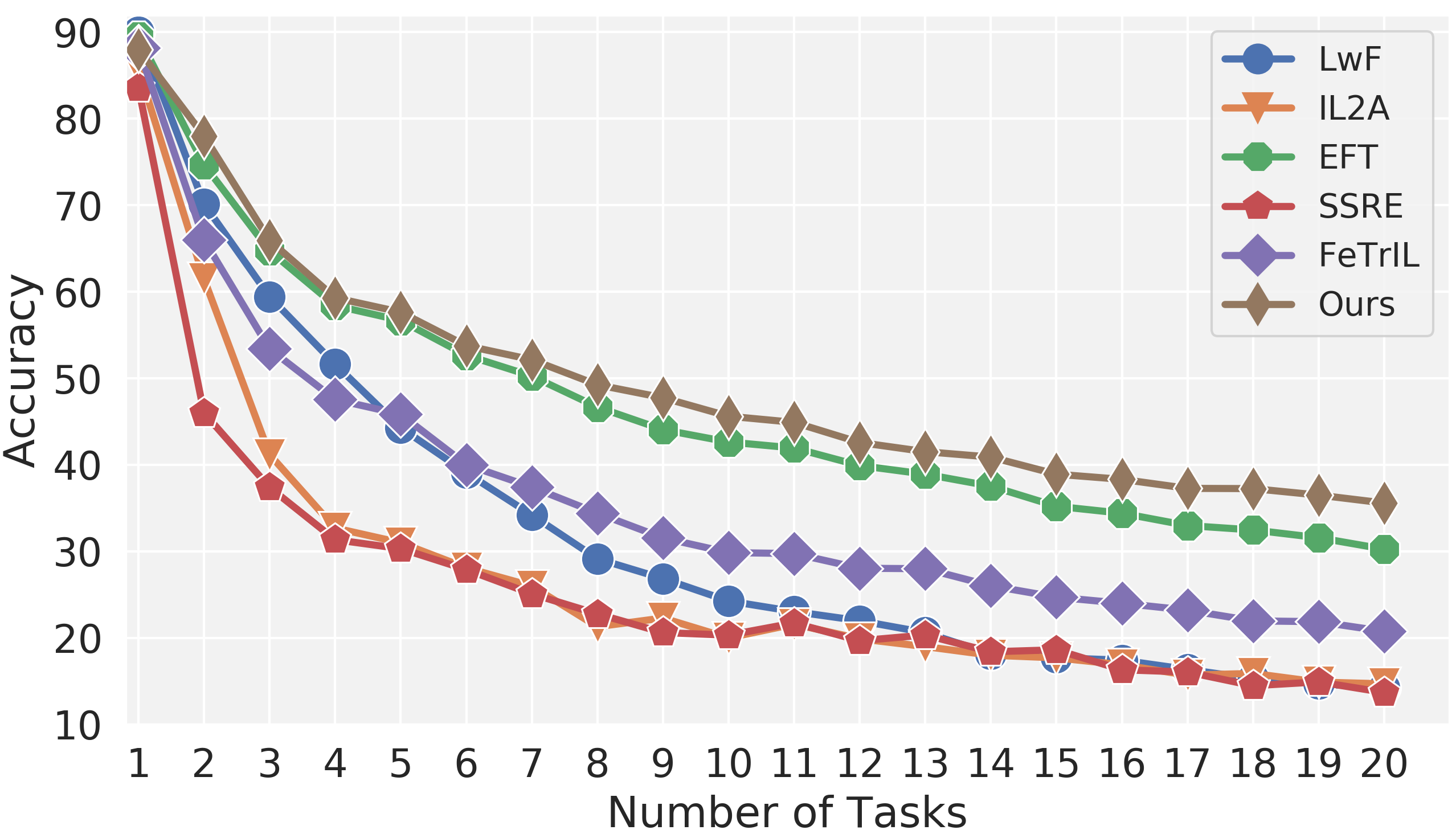}
	\end{subfigure}
	\vspace{-1em}
	\caption{Task-wise CIL results of SPG model on 5, 10 and 20 splits of CIFAR-100.}
	\label{fig:c100_20}
\end{figure*}
\begin{figure*}
	\setlength{\lineskip}{0pt}
	\centering
	\begin{subfigure}[H]{1\textwidth}
		\centering
		\includegraphics[width=0.8\textwidth]{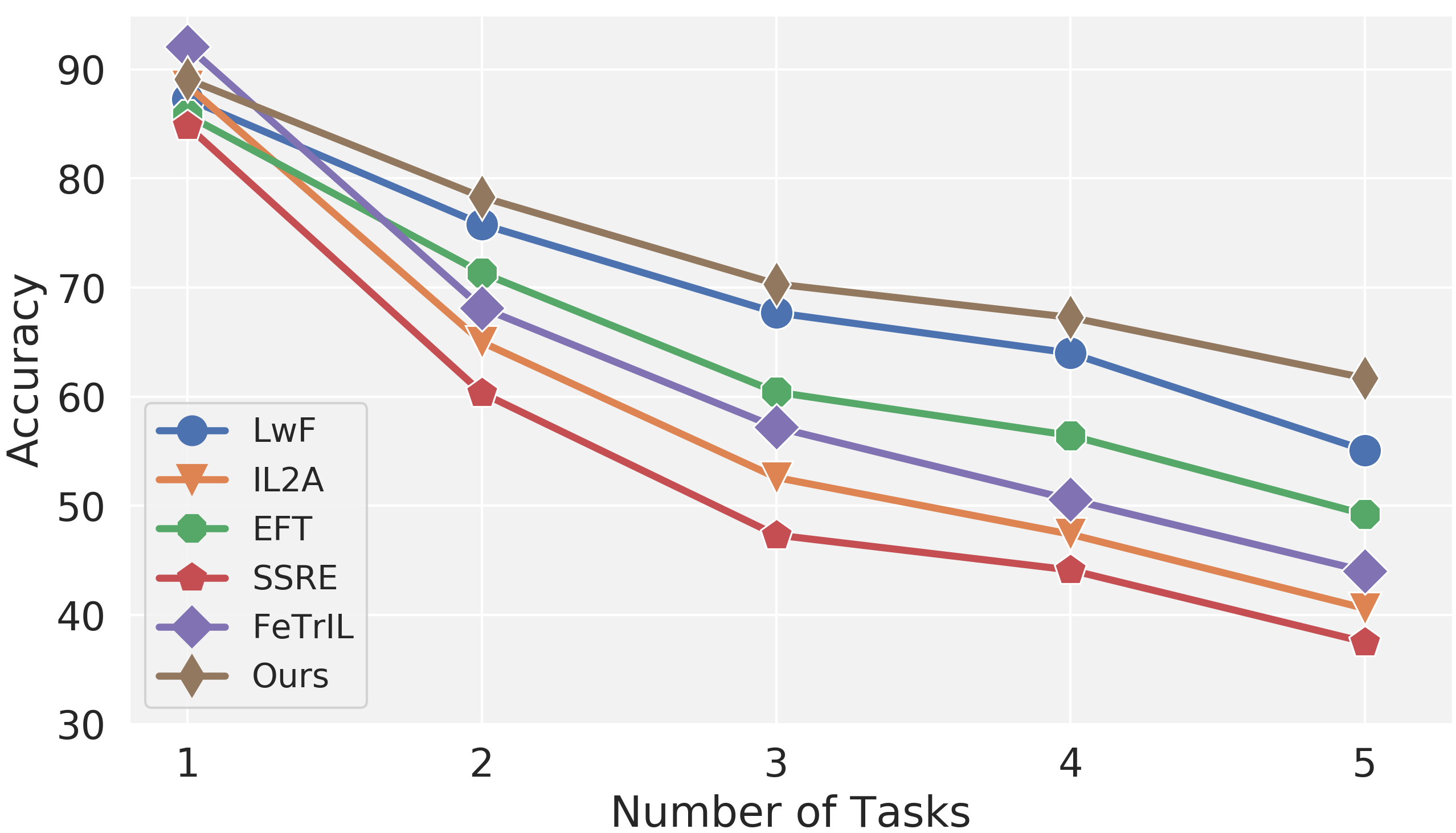}
	\end{subfigure}
	\begin{subfigure}[H]{1\textwidth}
		\centering
		\includegraphics[width=0.8\textwidth]{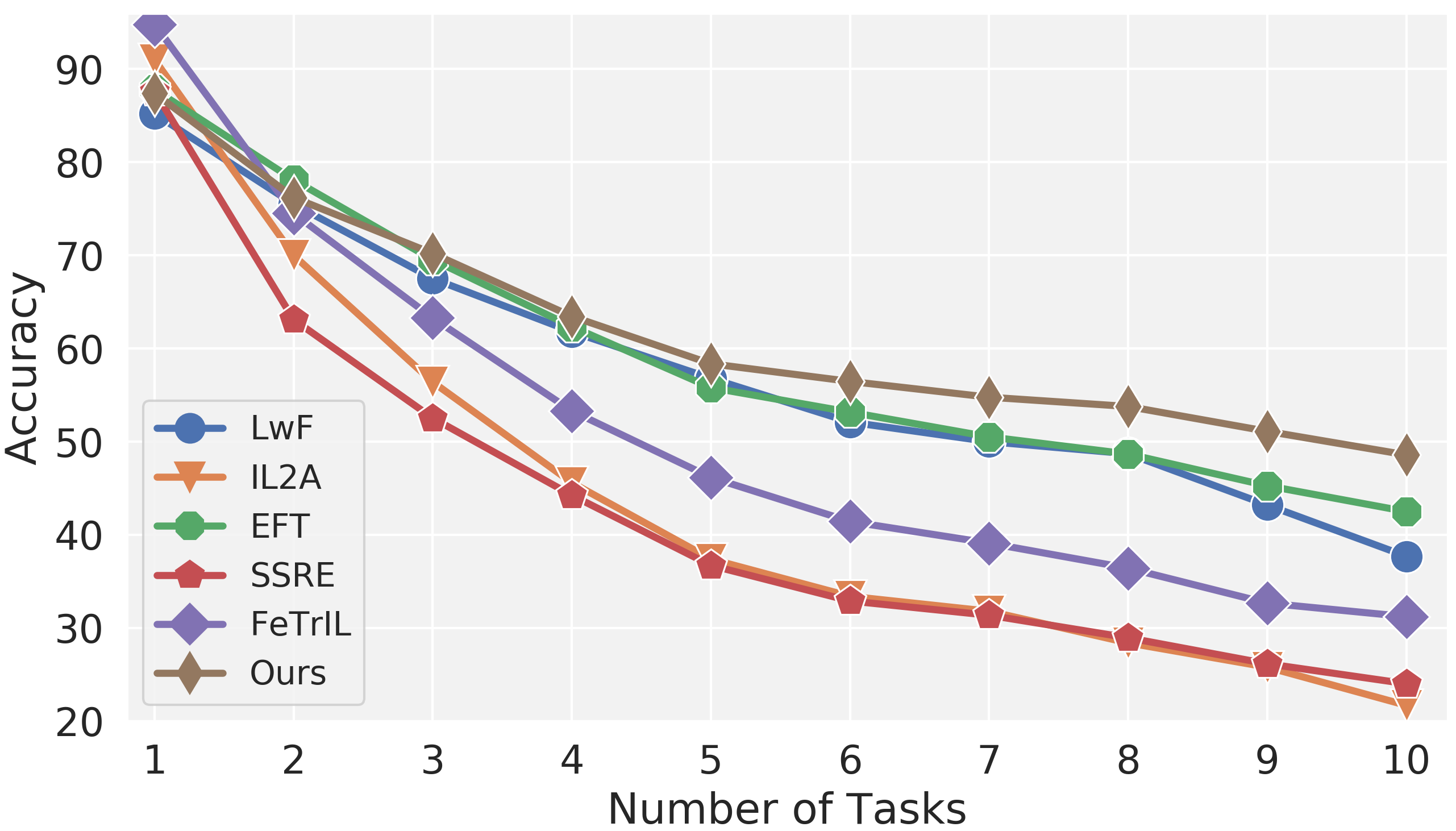}
	\end{subfigure}
	\begin{subfigure}[H]{1\textwidth}\ContinuedFloat
		\centering
		\includegraphics[width=0.8\textwidth]{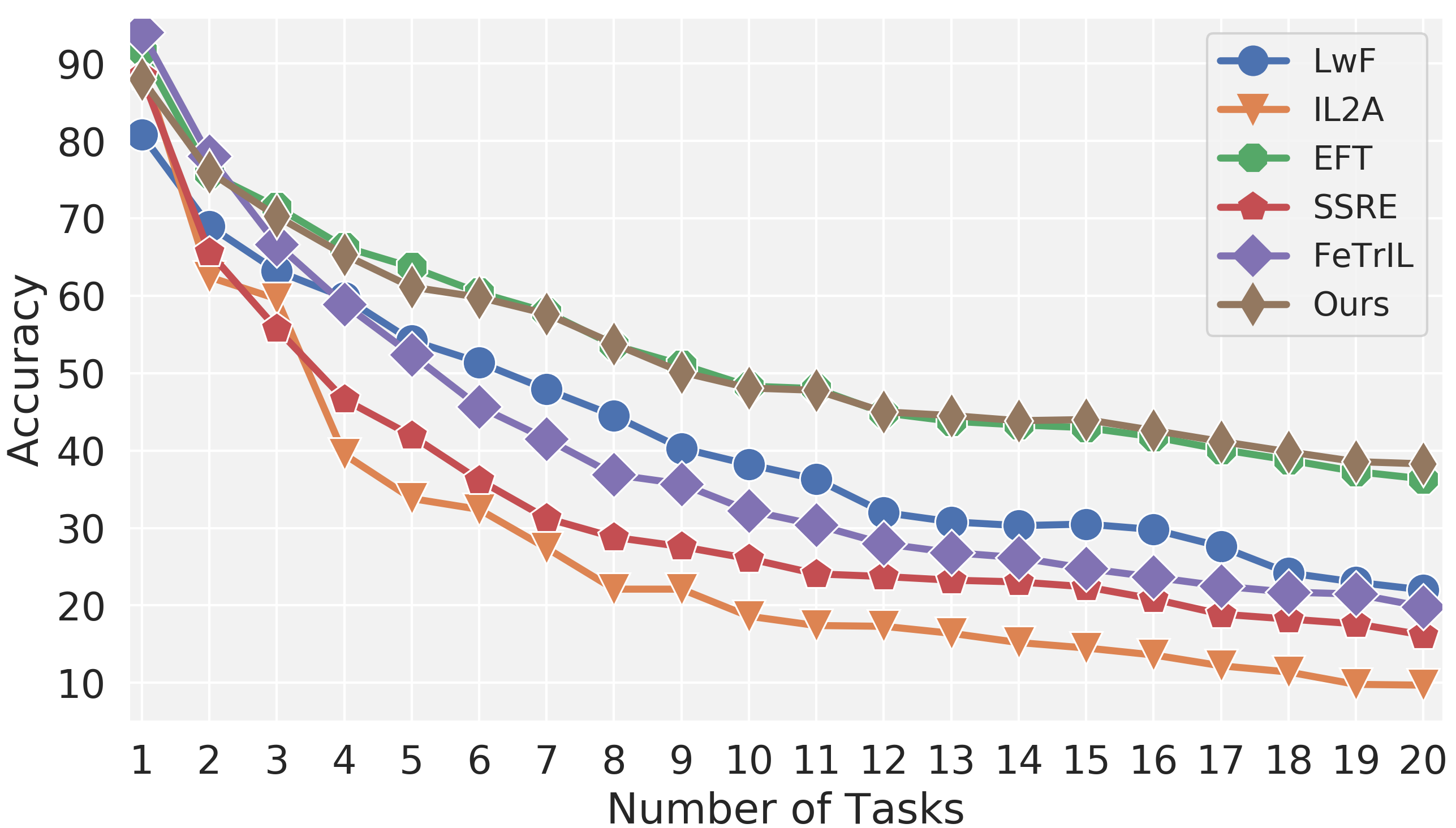} 
	\end{subfigure}
	\vspace{-1em}
	\caption{Task-wise CIL results of SPG model on 5, 10 and 20 splits of ImageNet-100.}
	\label{fig:im100_20}
\end{figure*}
\section{Experimental Settings}
In this section, we provide details about our hyperparameter settings and baselines.
\subsection{CIFAR-100}
\subsubsection{Training hyperparameters}
Since EFT~\cite{verma2021efficient} is our best performing baseline, we borrow the class order and hyperparameter settings (including seed) from their publicly available code. We train our model for $250$ epochs with batch size of $128$, initial learning rate of $0.01$, learning rate drop of $0.1$ at $100$, $150$ and $200$ epochs, SGD optimizer with momentum of $0.9$ and weight decay of $5e-3$.

We use ResNet-18~\cite{He2016} architecture for CIFAR datasets to evaluate our method. It should be noted that we train batch norm and linear layers from scratch for each task.
\subsubsection{Expansion hyperparameters}
We follow the same expansion hyperparameters for every task sequence. Let $\alpha_1$, $\alpha_2$, $\alpha_3$ and $\alpha_4$ be the number of filters used for creating the four residual blocks (using the $\_make\_layer()$ function in standard PyTorch implementation of ResNet-18 \footnote{github.com/pytorch/vision/blob/main/torchvision/models/resnet.py}). For each task, we increase the filters as follows:
\begin{equation}
	\label{eq:maxfilgrowth}
	\alpha_1 = \alpha_1 + 1,\ 
	\alpha_2 = \alpha_2 + 5,\ 
	\alpha_3 = \alpha_3 + 10,\ 
	\alpha_4 = \alpha_4 + 10 
\end{equation}
For the first task, $\alpha_1$ = $64$, $\alpha_2$ = $128$, $\alpha_3$ = $256$ and $\alpha_4$ = $512$ which is the standard ResNet-18 filter distribution. The criterion for selecting this hyperparameter is that we wanted to have an average parameter growth of around $4\%$ like EFT~\cite{verma2021efficient}.

For adaptive parameter growth, we define the minimum filter growth as:
\begin{equation}
	\label{eq:minfilgrowth}
	\alpha_1 = \alpha_1 + 1,\ 
	\alpha_2 = \alpha_2 + 1,\ 
	\alpha_3 = \alpha_3 + 1,\ 
	\alpha_4 = \alpha_4 + 1 
\end{equation}
The maximum filter growth is defined using Eq.~\ref{eq:maxfilgrowth}.
\subsubsection{Augmentations}
For class incremental learning (CIL), we apply $10$ instances of a random data augmentation scheme, along with the standard unaugmented test sample, to create the batch $\mathcal{X}_k$ (i.e., $A$ = $11$). It should be noted that we use the same random data augmentation scheme for task prediction that we use for training the network. Our data augmentation scheme is same as EFT~\cite{verma2021efficient}, i.e., from PyTorch library, we use:
\begin{enumerate}
	\item RandomCrop(32,\ padding=4)
	\item RandomHorizontalFlip() \item RandomRotation(10)
\end{enumerate}
\subsubsection{Baselines}
We borrow most of the baseline results from the EFT paper. We also run the publicly available code of IL2A~\cite{zhu2021class} using our class order, split (5/10/20) and seed setting. Results for SSRE~\cite{9878763} and FeTrIL~\cite{petit2023fetril} are obtained by running the PyCIL~\cite{zhou2021pycil} framework using our class order, split and seed settings. It should be noted that in the main paper, we define average incremental accuracy as the average accuracy for all seen classes.

\subsection{Tiny ImageNet}
\subsubsection{Training hyperparameters}
Like CIFAR-100, we borrow the class order, hyperparameter settings (including seed) and baselines from the EFT~\cite{verma2021efficient} paper. We train our model for $140$ epochs with batch size of $128$, initial learning rate of $0.01$, learning rate drop of $0.1$ at $70$, $100$ and $120$ epochs, SGD optimizer with momentum of $0.9$ and weight decay of $5e-4$. To evaluate our method, we use the VGG-16~\cite{simonyan2014very} architecture with batch norm for the Tiny ImageNet dataset.
\subsubsection{Expansion hyperparameters}
If $\alpha_{1,j}$ is the original number of filters for layer $j$ in VGG-16 and $\alpha_{i,j}$ are their values before task $i+1$, then for task $i+1$, we increase the filters as follows:
\begin{align*}
	\alpha_{i+1,j} = \alpha_{i,j} + 1\ if\ \alpha_{1,j}\ =\ 64\ or\ 128 \\
	\alpha_{i+1,j} = \alpha_{i,j} + 8\ if\ \alpha_{1,j}\ =\ 256\ or\ 512 \numberthis \label{eqn:tinyfilter}
\end{align*}
For adaptive parameter growth, we define the minimum filter growth as:
\begin{align*}
	\alpha_{i+1,j} = \alpha_{i,j} + 1
\end{align*}
We define the maximum filter growth using Eq.~\ref{eqn:tinyfilter}.
\subsection{ImageNet-100}
\subsubsection{Training hyperparameters}
We use the same class subset, class order and hyperparameter settings as DER~\cite{yan2021dynamically}. We train our model for $120$ epochs (unlike DER, we do not warm up) with batch size of $256$, initial learning rate of $0.1$, learning rate drop of $0.1$ at $30$, $60$, $80$ and $90$ epochs, SGD optimizer with momentum of $0.9$ and weight decay of $5e-4$.

We use ResNet-18~\cite{He2016} architecture for ImageNet dataset to evaluate our method. It should be noted that we train batch norm and linear layers from scratch for each task.
\subsubsection{Expansion hyperparameters}
We follow the same expansion hyperparameters as CIFAR-100, except for the ImageNet-100/20 split. If $\alpha_1$, $\alpha_2$, $\alpha_3$ and $\alpha_4$ are the number of filters used for creating the four residual blocks (using the $\_make\_layer$ function in standard PyTorch implementation of ResNet-18), then for each task in ImageNet-100/20 split, we increase the filters as follows:
\begin{equation}
	\label{eq:imagenetfilter}
	\alpha_1 = \alpha_1 + 2,\ 
	\alpha_2 = \alpha_2 + 10,\ 
	\alpha_3 = \alpha_3 + 10,\ 
	\alpha_4 = \alpha_4 + 10 
\end{equation}
For the first task, $\alpha_1$ = $64$, $\alpha_2$ = $128$, $\alpha_3$ = $256$ and $\alpha_4$ = $512$ which is the standard ResNet-18 filter distribution. This is because the ImageNet-100/20 split is harder than the corresponding CIFAR-100/20 split. For adaptive parameter growth and ImageNet-100/20 split, we define the minimum and maximum filter growths using Eq.~\ref{eq:minfilgrowth} and Eq.~\ref{eq:imagenetfilter} respectively.
\subsubsection{Augmentations}
For class incremental learning (CIL), we apply $20$ instances of a random data augmentation scheme, along with the standard unaugmented test sample, to create the batch $\mathcal{X}_k$ (i.e., $A$ = $21$). It should be noted that we use the same random data augmentation scheme for task prediction that we use for training the network. Our data augmentation scheme is same as~\cite{douillard2020podnet}, i.e., from PyTorch library, we use:
\begin{enumerate}
	\item RandomResizedCrop(224)
	\item RandomHorizontalFlip() \item ColorJitter(brightness=63 / 255)
\end{enumerate}
\subsubsection{Baselines}
We run the baselines LwF~\cite{li2017learning}, EFT~\cite{verma2021efficient} and IL2A~\cite{zhu2021class} using their publicly available code. Results for SSRE~\cite{9878763} and FeTrIL~\cite{petit2023fetril} are obtained by running the PyCIL~\cite{zhou2021pycil} framework. We use the same class subset, class order and seed for all our baseline experiments. It should be noted that in the main paper, we define average incremental accuracy as the average accuracy for all seen classes.
\subsection{Generative (GAN) Continual Learning}
We choose the StackGAN-v2~\cite{zhang2018stackgan} architecture for the incremental GAN experiment. StackGAN-v2 contains four blocks in the generator and discriminator networks. In the generator network, there are $1024, 512, 256, 128$ filters from first to the fourth block and the final image construction layer contains $64$ filters. We extend the last layer by $4$ filters; hence the respective increase in filters are $64, 32, 16, 8$ from first to the fourth block. During training of the $i^{th}$  task, all the previous task parameters are frozen; the parameter grows over the previous task parameters and not just over the global parameter. In our approach, we only grow the generator parameters and the discriminator is fixed for all the tasks; without any constraint, the discriminator parameter learns the current task. For the above discussed filter growth, the generator achieves a growth rate of $11.5\%$. We also observe that further filter growth shows better results. Our selected task sequences (cats, birds and churches) are highly diverse. The cat images are generally indoor or outdoor animal images; however, the next task (birds) are in a highly complex background and with fine-grained information; so the adaptation of birds from cats is difficult. Our model shows significant gains on the birds dataset using only $11.5\%$ extra parameters. The adaptation of churches from the birds dataset (birds to buildings) is also very difficult. Our proposed model adapts to this dataset and shows state-of-the-art results compared to the recent strong baselines.
\subsection{Heterogeneous Task Sequence}
We borrow the baselines and hyperparameter settings (including seed) from the EFT~\cite{verma2021efficient} paper. To evaluate our method, we use the VGG-16~\cite{simonyan2014very} architecture with batch norm.

\textbf{SVHN$\rightarrow$CIFAR10$\rightarrow$CIFAR100:} If $\alpha_{i,j}$ is the number of filters for layer $j$ in VGG-16 before task $i+1$, then we increase the filters as follows:
\begin{align*}
	&\alpha_{2,j} = \alpha_{1,j} + 10\\
	&\alpha_{3,j} = \alpha_{2,j} + 10\ if\ \alpha_{1,j}\ =\ 64\ or\ 128 \\
	&\alpha_{3,j} = \alpha_{2,j} + 20\ if\ \alpha_{1,j}\ =\ 256\ or\ 512
\end{align*}

\textbf{CIFAR100$\rightarrow$CIFAR10$\rightarrow$SVHN:} If $\alpha_{i,j}$ is the number of filters for layer $j$ in VGG-16 before task $i+1$, then we increase the filters as follows:
\begin{align*}
	&\alpha_{2,j} = \alpha_{1,j} + 10\ if\ \alpha_{1,j}\ =\ 64\ or\ 128 \\
	&\alpha_{2,j} = \alpha_{1,j} + 20\ if\ \alpha_{1,j}\ =\ 256\ or\ 512 \\
	&\alpha_{3,j} = \alpha_{2,j} + 10\\
\end{align*}
\subsection{Softwares} Experiments are run on a single V100 gpu using Linux, Python 3.6 and PyTorch 1.7.1 softwares.

\subsection{Input Processing} The data transformation scheme used in our method is borrowed from EFT~\cite{verma2021efficient} for CIFAR-100 and Tiny ImageNet datasets, while for ImageNet-100, we use the data transformation scheme used in~\cite{douillard2020podnet}. The codes for both these methods are publicly available.

\end{document}